\documentclass[letterpaper]{article} % DO NOT CHANGE THIS
\usepackage{aaai23}  % DO NOT CHANGE THIS
\usepackage{times}  % DO NOT CHANGE THIS
\usepackage{helvet}  % DO NOT CHANGE THIS
\usepackage{courier}  % DO NOT CHANGE THIS
\usepackage[hyphens]{url}  % DO NOT CHANGE THIS
\usepackage{graphicx} % DO NOT CHANGE THIS
\usepackage{natbib}  % DO NOT CHANGE THIS AND DO NOT ADD ANY OPTIONS TO IT
\usepackage{caption} % DO NOT CHANGE THIS AND DO NOT ADD ANY OPTIONS TO IT
\usepackage{algorithm}
\usepackage{algorithmic}

\usepackage{bm}
\usepackage{subfigure}
\usepackage{adjustbox}

\usepackage{newfloat}
\usepackage{listings}
\DeclareCaptionStyle{ruled}{labelfont=normalfont,labelsep=colon,strut=off} % DO NOT CHANGE THIS
\floatstyle{ruled}
\newfloat{listing}{tb}{lst}{}
\floatname{listing}{Listing}

\usepackage{amsmath,amssymb,amsfonts,amsthm}
\usepackage{multirow}

\nocopyright
\title{Towards Fully Automated Decision-Making Systems for Greenhouse Control: Challenges and Opportunities}
\author{
Yongshuai Liu, Taeyeong Choi, and Xin Liu
}
\affiliations{
University of California, Davis \\
\{yshliu, taechoi, xinliu\}@ucdavis.edu
}

\usepackage{bibentry}
\begin{document}

\maketitle

\begin{abstract}
Machine learning has been successful in building control policies to drive a complex system to desired states in various applications (e.g.~games, robotics, etc.). 
To be specific, a number of parameters of policy can be automatically optimized from the observations of environment to be able to generate a sequence of decisions leading to the best performance. 
In this survey paper, we particularly explore such policy-learning techniques for another unique, practical use-case scenario---\emph{farming}, in which critical decisions (e.g.,~water supply, heating, etc.) must be made in a timely manner to minimize risks (e.g.,~damage to plants) while maximizing the revenue (e.g.,~healthy crops) in the end.
We first provide a broad overview of latest studies on it to identify not only domain-specific challenges but opportunities with potential solutions, some of which are suggested as promising directions for future research.
Also, we then introduce our successful approach to being ranked second among $46$~teams at the ``$3$rd Autonomous Greenhouse Challenge'' to use this specific example to discuss the lessons learned about important considerations for design to create autonomous farm-management systems.
\end{abstract}

\section{Introduction}
With the rapid growth of global population, the demand for healthy and fresh food has been rising rapidly~\cite{FAO}. To improve sustainability of current food systems, novel AI and machine learning~(ML) techniques have been proposed for precision agriculture, where an individual-specific treatment is applied to each crop not only to maximize the total yields but also to utilize the available resources in an efficient manner~\cite{liu2023constrained}. 
To be specific, unhealthy crops can be identified with visual sensors~\cite{choi2022self}, spray nozzles can be automatically adjusted to only target weeds~\cite{salazar2022beyond}, the paths of mobile robotic sampler can be shortened for soil-quality monitoring~\cite{ruckin2022adaptive}, and water supply can also be determined for irrigation depending on some environmental factors~\cite{campoverde2021iot}.

\begin{figure}[t]
    \centering
    \includegraphics[width=0.8\linewidth]{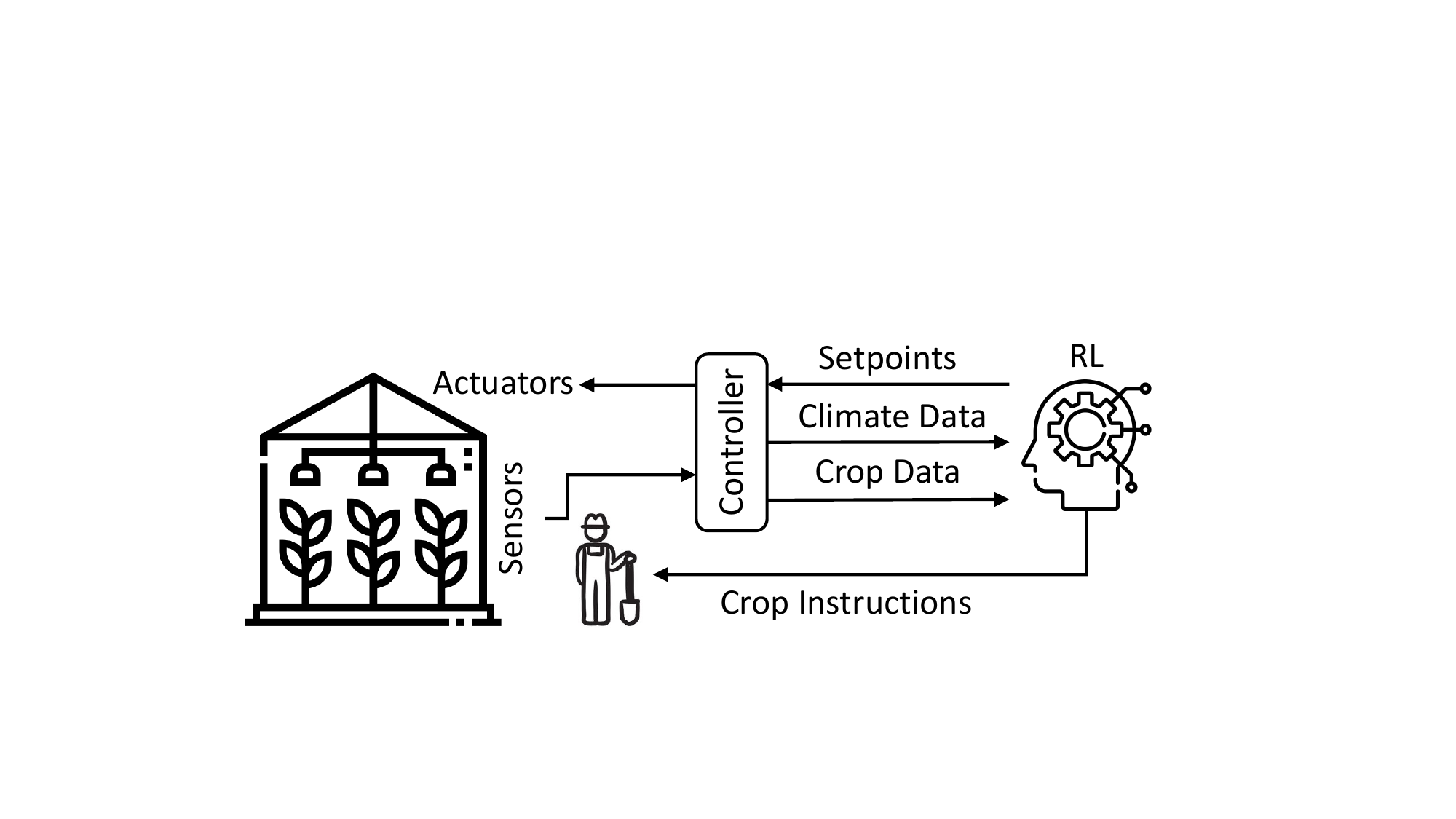}
    \caption{Illustration of greenhouse control system where a centralized controller autonomously utilize the actuators and human to maximize productivity based on various types of data from sensors.}
    \label{fig:illustration}
    \vspace{-0.3cm}
\end{figure}

In this survey paper, we consider the problems of learning control policies for automated decision-making in \emph{farming}, particularly inspired by a great success of reinforcement learning~(RL)~\cite{liu2024towards} in other domains---such as computer games~\cite{vinyals2019grandmaster}, robotics~~\cite{agostinelli2019solving}, self-driving cars~\cite{liang2018cirl}, and recommender systems~\cite{afsar2021reinforcement}.
In this paper, however, our focus is not on micro-level motion control or perception capabilities of robotic devices but instead on generating higher-level policies as autonomous decision makers for \emph{farm management} that could command separate actuators or agents to maintain the best conditions of environment for maximizing the final profit. 

To be specific, Fig.~\ref{fig:illustration} shows farm operation as a closed-loop control system that could repeatedly determine the parameters of relevant actuators to adjust physical properties (e.g.,~temperature, humidity, and lighting) based on the latest sensor readings of the climate and plants. 
This decision problem may become more complex since as productivity of crop is maximized, the expenses of resources must also be considered.
For example, some specialized sensors could be installed in the soil for the control system to automatically decide to gate in only an optimal amount of chemical fertilizers or water to improve soil quality of a particular region. 

\begin{table*}[t] % \small%
\centering
\begin{adjustbox}{width=\textwidth}
% \begin{tabular}{|p{1.1cm}|p{2.3cm}|p{2cm}|p{3.2cm}|p{1.5cm}| p{1.1cm}|}
\begin{tabular}{|c|c|c|c|c|c|}
\hline
Task & Publication & Representative State Variables & Control Variables & Objectives & Methods \\
\hline\hline
\multirow{4}{*}{GC} & {\cite{cao2022igrow}} & Out. weather, In. TMP/HUM/$\text{CO}_2$, Crop prop. & TMP, $\text{CO}_2$, Light, Irrig. & $\uparrow$ yield, $\downarrow$ cost & GA, RL \\
\cline{2-6}
 & \cite{ajagekar2022deep} & In. aerial/floor/wall/ceiling TMP & In. TMP & $a<\text{T}<b$,  $\downarrow$ cost & RL \\
\cline{2-6}
 & \cite{wang2020deep} & Out./In. TMP \& HUM, $\text{CO}_2$, Light, Vent., Irrig., Drain. & In. TMP/HUM/$\text{CO}_2$ & $\uparrow$ yield, $\downarrow$ cost & RL, IL  \\
% \cline{2-6}
 % & \cite{hemming2020cherry} & Data 2 & No \\
\cline{2-6}
 & \cite{van2015optimal} & In. TMP/HUM/$\text{CO}_2$ & TMP, Vent., $\text{CO}_2$ & $a<\text{T, H, C}<b$, $\downarrow$ cost &  \\
\hline
\multirow{3}{*}{NM}  & \cite{wu2022optimizing} & Amount of N, Out. weather, Crop prop., Soil moist. & N app. & $\uparrow$ yield, $\downarrow$ cost, $\downarrow$ N leaching & RL \\
\cline{2-6}
 & \cite{overweg2021cropgym} & Out. weather, Crop/Soil prop., Irrig. & N app. & $\uparrow$ yield, $\downarrow$ cost & RL \\
\cline{2-6}
 & \cite{garcia1999use} & Amount of N, Crop prop., Sowing time  & Seed rate, N~app. & $\uparrow$ yield, $\text{N}<\tau$ & RL \\
 \hline
 NM \& IC & \cite{tao2022optimizing} & Amount of N, Out. weather, Crop prop., Soil moist. & N app., Irrig. & $\uparrow$ yield, $\downarrow$ cost, $\downarrow$ N leaching & RL, IL \\ 
 \hline
  \multirow{3}{*}{IC}   & \cite{campoverde2021iot} & Soil moist. & Irrig. & $a<\text{S}<b$ & RL\\
\cline{2-6}
 & \cite{sun2017reinforcement} & Soil moist. & Irrig. & $\uparrow$ yield, $\downarrow$ cost & RL\\
\cline{2-6}
 & \cite{bergez2001comparison} & Soil moist., TMP & Irrig. & $\uparrow$ yield, $\downarrow$ cost & DP, RL \\
 \hline
\multirow{2}{*}{CR} & \cite{jena2022environmental} & Out. TMP/HUM/weather, Soil~pH & $23$~crop selections & $\uparrow$ yield & BO, SL \\
\cline{2-6}
 & \cite{kirsch2019improving} & Out. weather, Water avail., Crop prop. & Wheat varieties & $\uparrow$ yield & MB \\
 \hline
 \multirow{1}{*}{PS} & \cite{baudry2021optimal} & N/A & Four possible dates & $\uparrow$ yield & MB \\
 \hline
\end{tabular}
\end{adjustbox}
\caption{Previous works on autonomation of farm management; GC:~Greenhouse Control, NM:~Nitrogen Management, IC:~Irrigation Control, CR:~Crop Recommendation, PS:~Planting Scheduling; TMP~(T):~Temperature, HUD~(H):~Humidity, Crop prop.:~Crop property, Irrig.:~Irrigation., Drain.:~Drainage, N:~Nitrogen; $a$ and $b$ are the certain lower and upper bound, respectively.}
\label{tab:list_of_work}
% \vspace{-0.4cm}
\end{table*}

This paper aims to offer a broad overview of latest studies on AI~methodology for farm-management automation in various scenarios---e.g.,~greenhouse control~\cite{cao2022igrow,ajagekar2022deep,wang2020deep,van2015optimal}, irrigation~\cite{campoverde2021iot,bergez2001comparison,navarro2016decision}, crop recommendation~\cite{kirschner2019stochastic,jena2022environmental}, planting scheduling~\cite{baudry2021optimal}, and nitrogen management~\cite{overweg2021cropgym,wu2022optimizing,tao2022optimizing,garcia1999use}.
Also, we discuss domain-specific challenges and potential opportunities to consider in future AI~research for farm automation. 

Moreover, we introduce our investigations on model-based RL and Bayesian Optimization~(BO) to be ranked second among $46$ teams at the ``$3$rd International Greenhouse Challenge'', in which the goal was to optimize the control of various parameters of greenhouse facilities on simulation to maximize the expected \emph{net profit} from grown lettuces.
This specific instance of study is used to discuss the lessons learned particularly about crucial design choices to construct successful autonomous farm-management systems. 
% To the best of our knowledge, we are the first team from the competition that reports specific methodologies adopted to provide useful guidance not only for the scientists in relevant fields but also for the AI~practitioners who would like to start working on agricultural problems.
% for the purpose of  with the lessons learned for further research.

Recently, \cite{gautron2022reinforcement} provided a survey of the studies on crop-management support systems. Yet, their scope was limited to RL~methods only with general perspectives of expected challenges, whereas ours covers a broader selection of AI~algorithms along with our practical case study.

\section{Background}
\label{sec:background}

% \subsection{Relevant Tasks}

In this section, we examine previous works that have been conducted to automate decision making processes for farming. 
As shown in Table~\ref{tab:list_of_work}, Greenhouse Control~(GC), Nitrogen Management~(NM), Irrigation Control~(IC), Crop Recommendation~(CR), and Planting Scheduling~(PS) have been most actively studied in this field of research. 
In particular, proposed methods have been developed to maximize the ``net profit'' by increasing the overall yield whilst using the minimal amounts of resources (e.g.,~electricity, gas, water, fertilizers, etc.)~\cite{cao2022igrow,wang2020deep,overweg2021cropgym,sun2017reinforcement,bergez2001comparison}.
Still, some approaches have been designed to also reduce potential negative impacts on the natural environment (e.g.,~nitrate leaching)~\cite{wu2022optimizing,tao2022optimizing,garcia1999use}.

In addition, state and action sets are generally composed of various variables to encode environmental conditions (e.g.,~climate, $\text{CO}_2$ concentration, and soil moisture) or mechanical actuation (e.g.,~ventilation, lighting, irrigation, and drainage). 
Also, some tasks such as CR and~PS consider the discrete decisions---i.e.,~variety selection or planting date---to be made only once~\cite{jena2022environmental,kirsch2019improving,jena2022environmental} while other scenarios may involve more frequent controls for climate adjustment or irrigation on an hourly or daily basis~\cite{cao2022igrow,wang2020deep}.

Most authors utilized data-driven approaches to produce decision-making policies from historical or simulated data~\cite{kulkarni2018predictive,tantalaki2019data}. 
To be specific, RL~has appeared as a significantly useful paradigm of learning~\cite{overweg2021cropgym,campoverde2021iot}, and its variants such as imitation learning~(IL) and multi-armed bandit~(MB) have also been considerably successful~\cite{wang2020deep,tao2022optimizing,kirsch2019improving,baudry2021optimal}. 
For sample efficiency, \cite{jena2022environmental}, however, adopted a BO~algorithm in optimizing a SVM~classifier. 
More in-depth discussions on each method follow through Sec.~\ref{sec:challenges} and Sec.~\ref{sec:opportunities}.  

\section{Autonomous Greenhouse Control}~\label{sec:greenhouse}
In this section, we introduce our example of use case to produce an autonomous agent for farm operation in the $3$rd Autonomous Greenhouse Challenge\footnote{\url{https://www.wur.nl/en/article/online-challenge.htm}}.
In particular, our team---so called ``AggieAI''---was placed second among $46$~groups on Part~B Machine Learning Challenge (cf. Table~\ref{tab:scores}), in which the goal was to optimize the control of various parameters of simulated greenhouse facilities to maximize the expected profit from well-grown lettuces. 
% To the best of our knowledge, we are the first team from the competition that shares specific attempts of methodologies to relate to promising directions of future studies. 

% Simulation, Input/Output
\subsection{Lettuce Growth Simulator}
\label{sec:simulator}

Given the slow and vulnerable nature of agricultural systems that require protection from potential damage, the use of a realistic simulator becomes crucial for optimizing control policies based on data. Instead of relying solely on real-world agricultural systems, utilizing a simulator can offer several advantages.

In the specific context of the competition, participants were provided access to a web-based simulator that replicated greenhouse environments. This simulator generated climate and crop development-related variables (e.g., fresh weight, dry matter content) within the greenhouse in response to simulated outside weather conditions and the decisions submitted by participants. These decisions included setpoints for heating, ventilation, lighting, and CO2 supply systems. Additionally, participants had the opportunity to adjust the crop density to maximize space utilization, which could be done less frequently compared to hourly actions.

Each growing cycle within the simulator also evaluated the quality of the control policy using a predefined formula for net profit. This formula took into account the productivity of lettuce and the total maintenance costs. Consequently, an effective policy should not focus solely on maintaining high temperatures to achieve larger crops, as the associated heating expenses could significantly reduce the final profit.

The competition provided four simulators, each with distinct conditions related to weather, greenhouse type, and lettuce variety. During the preparation phase, participants could interact with the simulator to develop their algorithms using limited samples. In the Online Challenge phase, the developed algorithm needed to be adaptable to control the growth of a virtual crop under changing conditions and within limited time constraints.

For further details regarding the simulator, we recommend referring to the work by~\cite{Zwart2021}.

Overall, the use of a realistic simulator in the competition offered participants the opportunity to optimize control policies based on simulated data. This approach allowed for experimentation and algorithm development while considering various environmental factors and cost implications. By utilizing the simulator, participants could devise strategies that balanced crop productivity with maintenance costs, ultimately aiming to maximize net profit.

\subsection{Farm Management as a Constrained MDP}
\label{sec:mdp}

As in~\cite{cao2022igrow}, the farm-management problem can be more formally formulated as a Markov Decision Process~(MDP) 
with tuple $(S,A,R,P,\gamma)$. %~\cite{liu2021clara}. 
To be specific, the observations (i.e.,~temperature, images of crop, etc.) from deployed sensors at each time~$t$ constitute the state $s_t \in S$ of the whole system. 
In addition, the action set~$A$ consists of possible joint actions~$a_t$ (e.g.,~open air inlet/supply water) for accessible actuators (e.g.,~ventilation/irrigation systems) to influence certain properties (e.g.,~air quality/soil moisture level). 
Also, the reward function~$R: S \times A \times S \mapsto \mathbb{R}$ may be dependent upon the taken action~$a_t$ under state~$s_t$ and its resulting sate~$s_{t+1}$. For example, $r_t$ could be defined as $+1$, if $s_{t+1}$ ensures a higher yield than $s_{t}$.

Furthermore, some cost functions~$C: S \times A \times S \mapsto \mathbb{R}$ may also be predefined as suggested in~\cite{liu2021policy,liu2020constrained} to take into account $1$)~``cumulative'' constraints (e.g.,~total usage of fertilizer) and $2$)~``instantaneous'' ones (e.g.,~minimum temperature requirement).  
Moreover, the state transition function~$P: S \times A \times S \mapsto [0, 1]$ represents~$\Pr(s_{t+1} | s_t,a_t)$, which governs the stochastic dynamics of the world, albeit unknown in most realistic scenarios. 
Lastly, $\gamma$~is the discount factor, possibly defined to be different for reward and cost calculations. 

% In particular, as suggested in~\cite{liu2021policy}, we also consider two types of constraints:
% % , e.g., resource limits, fairness, and other network constraints. 
% 1)~\emph{Cumulative} constraints require the accumulated costs to be within a certain limit (e.g.,~the total usage of water) and 2)~\emph{Instantaneous} ones require each action/state not to violate any pre-defined condition at each time slot---e.g.,~ minimal humidity. 

Thus, control-policy learning is to obtain the optimal policy~$\pi^*$ that can maximize the discounted cumulative reward
% $$
    $J_{R}^{\pi}= \mathbb{E}_{\tau \sim \pi}[\sum_{t=0}^{\infty}\mathit{\gamma}^{t}\mathit{R}(s_{t},a_{t},s_{t+1})]$
% $$
while satisfying both discounted cumulative constraints
% $$
    $J_{C_{i}}^{\pi}=\mathbb{E}_{\tau \sim \pi}[\sum_{t=0}^{\infty}\gamma^{t}\mathit{C_i}(s_{t},a_{t},s_{t+1})]$
% $$
and instantaneous constraints~$C_j$, where $\tau = (s_{0}, a_{0},s_{1}, a_{1}... )$ is a trajectory.
That is, the final form for optimization can be defined as:  
\begin{align}
& \underset{\pi}{\text{maximize}} & &\max_{\pi} J_{R}^{\pi}\label{eq:objective}  \\
&\text{subject to}  & & J_{C_{i}}^{\pi} \leq \omega_i,  \text{for each $i$}, \label{eq:constraint}  \\
& & & C_{j}(s_t, a_t)\leq \epsilon_j, \text{for each $j$ and $t$}. \label{eq:instconstraint} 
\end{align}

An extension to Partially Observable Markov Decision Process~(POMDP) could be considered to incorporate the concept that sensors can only capture partial information of states~\cite{gautron2022reinforcement}. Yet, the detailed formalization is beyond the scope of this paper.
\section{Practical Challenges}
\label{sec:challenges}
% Although a few methods have used RL in autonomous farming control, most of them are in the realm of simulation. Challenges exist to apply RL in the real-world system. We summarize the challenges here and provide potential technical solutions in Sec.~\ref{sec:opportunities}, as shown in Table~\ref{tab:co}.
In this section, we describe challenges in developing autonomous farm operators. 
In particular, domain-specific natures of agricultural systems are linked to potential technical problems with related terms in~ML to effectively bridge between readers from both fields. 
% Despite the great success of RL in various applications, unique technical challenges exist in agricultural domains. 
% We first describe five primary ones in this section and discuss potential solutions in the following section (cf.~Table~\ref{tab:co}). 

\subsection{Complexity of Agricultural Systems}
\label{sec:complex}

Farming is inherently a highly complex activity due to the need to understand and manage the relationships among various environmental factors. These factors include plants, soil, fertilizers, climates, and more, all of which interact with each other dynamically over time. Even under carefully controlled conditions, the same species of crop can exhibit diverse phenotypes, leading to different interactions with the environment.

When attempting to model and optimize farming systems, it becomes impractical to define a state space that encompasses all the fine-grained information about individuals and environments. The sheer size of such a state space would make it infeasible to learn and evaluate every possible state-action trajectory. As a result, a trained controller would have to rely on interpolation or extrapolation techniques to navigate unseen states when deployed, which introduces the risk of poor generalizability.

Similarly, relying solely on physics-based simulations to guide farming practices may not capture all the intricacies of complex environmental phenomena. This limitation can result in a simulation-to-reality gap, where the parameters of a control policy obtained from simulations are likely to be suboptimal and may fail to perform effectively in real-world scenarios.

Therefore, both the limitations of managing an enormous state space and the simulation-to-reality gap highlight the challenges faced in farming optimization. It is crucial to develop approaches that strike a balance between capturing the complexity of farming systems and ensuring practical applicability in real-world conditions.

% be easily built as interactive testbeds for RL, which also hinders active research. 
% The control parameters for different actuators may be in different \emph{temporal scale}, such as the temperature could be set up in each hour, while some parameters can only be set up once before the planting, e.g., the CO2 supply rate, the light intensity and the material of the greenhouse screen, where RL cannot work efficiently.
% The complexity of the agriculture system also causes the absence of a realistic benchmark simulator/data.
% To speed up the learning process and reduce the required data, the human experts are required in some cases, as discussed in Sec~\ref{sec:human}. For example, in the greenhouse control task, the expert knowledge that the temperature increases with the CO2 dosing significantly reduces the action space.

\subsection{Resource Constraints \& Safety Requirements}
\label{sec:constraints}
% Although the machine-learning based controls work better than human, the SOTA is that the control agent is almost a black box. So it's inevitable to cause the trustworthiness problem. The trustworthiness can be divided to two parts: safety and explain-ability.

The primary objective of smart farming extends beyond improving productivity; it aims to save valuable resources such as water, electricity, and human labor. This is achieved by executing actions that are optimized based on relevant objective functions. In addition to resource efficiency, it is crucial to consider the potential risks associated with certain actions in specific states, as they could lead to irreversible outcomes such as crop damage or facility destruction. This becomes especially important when learning is conducted directly in physical farming environments.

To address these challenges, a concept known as ``constrained optimization" must be incorporated into the farming system. Constrained optimization involves setting up rules and limitations based on domain knowledge to ensure that essential requirements and safety measures are met. By considering these constraints, the system can strike a balance between achieving optimal objectives and avoiding undesirable outcomes.

It is worth noting that smart farming scenarios differ significantly from simpler contexts like card games such as Dou Dizhu, where players typically focus on a single objective of gaining more points to win. In contrast, smart farming requires a more comprehensive approach that accounts for multiple objectives, constraints, and domain-specific considerations to optimize resource utilization and maintain sustainable agricultural practices.
\subsection{Slow Growth Rates}
\label{sec:slow}

In comparison to mechanical systems, the development of a plant's state occurs at a considerably slower pace. For instance, the formation of apples begins in spring but progresses throughout summer and fall as they grow larger and ripen. This characteristic of plant growth introduces challenges in data collection processes within the agricultural domain, often resulting in ``data scarcity" during relevant research endeavors.

Furthermore, in agriculture, decisions made by control policies, such as heating adjustments, do not yield immediate outcomes. The effects of these decisions, such as increased crop yield, may only become observable after days or even months. This phenomenon presents a well-known challenge known as the ``delayed reward" problem in reinforcement learning (RL) for tasks with long time horizons (Pathak et al., year). We consider farming to fall within this task family, as decisions made today can have long-lasting consequences that may not be apparent immediately.

The combination of slow plant development and delayed rewards underscores the importance of carefully designing and evaluating control policies in agricultural systems. It highlights the need for patience, as well as the ability to anticipate and account for long-term effects when making decisions related to farming practices. Additionally, addressing the delayed reward problem in RL approaches for agricultural management becomes crucial for developing effective and sustainable control policies.

In summary, the slow rate of plant development and the delayed reward nature of farming tasks pose significant challenges in data collection and decision-making processes. Acknowledging and addressing these challenges is essential for successful research and the development of control policies that maximize agricultural productivity and sustainability.

\subsection{Interaction with Human Farmers}
\label{sec:interact_with_human}

While farming has witnessed a transition towards full automation, it remains crucial that every decision made in agricultural management is human-centered. These decisions often involve human labor, as well as the production of products for stakeholders and consumers. Therefore, AI systems employed in farm management must possess a level of ``explainability" that allows them to present logical causality behind their decisions. This capability is vital to establish trust and ensure that the outcomes generated by the system are both productive and safe.

Without explainability, granting full autonomy to AI systems in farm management would be challenging, as human users would be hesitant to fully trust the decisions made by the system. The ability to understand the reasoning and rationale behind the AI system's choices is essential for building confidence and facilitating effective collaboration between humans and machines.

Furthermore, an AI decision-maker in farm management should have the capability to continually learn from human experts, augmenting and adapting its knowledge base over time. Even in situations where immediate access to relevant data samples is limited, the system should be able to update its internal knowledge graph with new scientific findings and insights. This ensures that the AI system remains up-to-date and aligned with the latest advancements in agricultural research and practices.

Additionally, effective utilization of multimodal data is crucial for an AI decision-maker in farming. This encompasses integrating and leveraging various types of data, ranging from human knowledge represented in mathematical logic or natural languages to numerical sensor readings. Specialized learning strategies may be required to effectively extract insights from these diverse data sources, enabling the AI system to make informed decisions based on a comprehensive understanding of the agricultural domain.

In summary, human-centered decision-making remains essential in farming, even as automation becomes more prevalent. AI systems for farm management must possess explainability, providing logical causality for their decisions to establish trust and ensure productivity and safety. Continual learning from human experts and the effective utilization of multimodal data further enhance the capabilities of AI decision-makers in the agricultural domain. By incorporating these aspects, AI systems can serve as valuable tools that collaborate with human stakeholders to optimize farm operations and achieve sustainable and productive outcomes.

\begin{table}[t]
% \scriptsize
\centering
\caption{Challenges and Opportunities}
\label{tab:co}
\begin{adjustbox}{width=\columnwidth}
\begin{tabular}{l|cccc}
\hline
& System & Constraints & Slow & Human  \\
&Complexity&\&Safety&Growth&Interaction\\
\hline
% Model predictive control& Multi control variables, constraints &Computation time and cost \\
BO & \checkmark & & \checkmark & \\

Novelty-driven& \checkmark & & \checkmark &\\

Constrained RL&  & \checkmark&  &\\

Demonstrations &  &\checkmark & \checkmark & \\

Reward shaping&  & &\checkmark  &\\

MBRL&  & & \checkmark &\\

Sim2real & \checkmark & & \checkmark & \\

Meta Learning & \checkmark & & \checkmark &\\

Transfer Learning & \checkmark & & \checkmark &\\
% Inverse RL & Data-driven rewards & Non-optimal rewards\\

Explainable RL&  & &  &\checkmark\\

Multi-model ML&  & &  &\checkmark\\

% Inverse RL&  Intermediate rewards, Generalization & Computation complex  \\
% Explainable RL& &\\
\hline
\end{tabular}
\end{adjustbox}
% \vspace{-0.1cm}
\end{table}

\section{Potential Opportunities}
\label{sec:opportunities}

This section discusses useful methods to potentially address the aforementioned challenges when developing autonomous agents for decision-making for farming~(Table~\ref{tab:co}). Moreover, Table~\ref{tab:pc} summarizes the advantages and limitations of each method. 

\subsection{Informative Sampling}

As stated in Sec.~\ref{sec:complex} and Sec.~\ref{sec:slow}, agricultural systems are typically extremely ``complex'' and progress at a ``slow'' speed, so gathering sufficiently large data can be significantly challenging in developing an automated farm. 
Here, we thus present two promising approaches to create a dataset that can be limited in size but significantly high-quality in terms of informativeness.
% prioritize high-quality observations that would be information-rich during data collection. 

\subsubsection{Bayesian Optimization}~\label{sec:bo}
% As the control parameters may be in different temporal scales (cf. Sec.~\ref{sec:complex}), RL cannot make sure to work efficiently on all of them.
% Considering some of the challenges above, 
We highly recommend the use of Bayesian Optimization (BO) as a valuable method, as it has demonstrated success in efficiently searching for optimal parameters with only a few evaluations of a candidate set of parameters. This effectiveness has been documented in studies such as the tutorial by Frazier et al. (2018)~\cite{frazier2018tutorial}.

\begin{figure}
    \centering
    \includegraphics[width=0.7\linewidth]{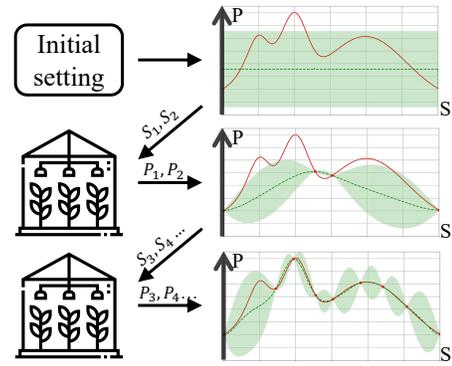}
    \caption{Workflow with BO, in which P and S represent the net profit and the control setpoints, respectively.}
    % After training in a controlled environment, the learned model can be used to provide optimal treatment recommendations in the ﬁeld.}
    \label{fig:bo}
    % \vspace{-0.3cm}
\end{figure}

BO is particularly advantageous because it leverages previous observations to systematically plan the selection of informative parameter sets for evaluation. By incrementally modeling the relationship between each parameter and the objective, such as maximizing net profit in farming systems (as illustrated in Fig.~\ref{fig:bo}), BO can guide the search process effectively. Through carefully planned and meaningful evaluations, BO enables the discovery of the most desirable parameters in an efficient manner. This is especially valuable in the context of farming, where systems are highly complex and involve numerous controllable parameters, as discussed in Section~\ref{sec:complex}. Additionally, observational data in farming systems can be scarce due to the slow changes that occur in the system over time, as mentioned in Section \ref{sec:slow}.

An example of the successful adoption of BO in agriculture can be found in the work of Jena et al. (2022), where they applied a BO technique for crop recommendation~\cite{jena2022environmental}. In our own case studies, presented in Section \ref{sec:greenhouse}, we also showcase a specific instance of solving greenhouse control problems using a BO algorithm.

Overall, BO offers a promising approach for optimizing farming systems by efficiently searching for the most favorable parameters, even in the face of complexity, limited data, and slow system changes.

\begin{table*}[t]
% \small
\centering
\caption{Pros and cons of different methods}
\label{tab:pc}
\begin{adjustbox}{width=\textwidth}
\begin{tabular}{l|ll}
\hline
Methods  & Pros & Cons  \\
\hline
% Model predictive control& Multi control variables, constraints &Computation time and cost \\
BO & Data-efficiency  & Low-dimensional parameters \\

Novelty-driven& Data-efficiency, Escape local optima & May explore unsafe actions\\

\multirow{2}{*}{Constrained RL} & \multirow{2}{*}{Resource and safety constrained controls} & Data inefficiency with cumulative constraints \\
&&Computation complex\\

Demonstrations & Data-efficiency & Need human efforts, Less generalization \\

Reward shaping&  Construct dense reward & reward cumulative error  \\

\multirow{2}{*}{Model-based RL}& \multirow{2}{*}{Data-efficiency, Time-efficiency} & Need well-learned model\\
& & Computation complex to learn models\\

Sim2real & Data-efficiency, Time-efficiency & Need well-learned simulators \\

Meta Learning\&TL & Generalization, Adapt to new environment & Data inefficiency during training\\
% Inverse RL & Data-driven rewards & Non-optimal rewards\\

Multi-model ML& Multi-model data, Information complementary & Need specialized approach to align modalities\\

% Inverse RL&  Intermediate rewards, Generalization & Computation complex  \\
% Explainable RL& &\\
\hline
\end{tabular}
% \vspace{-0.2cm}
\end{adjustbox}
\end{table*}

\subsubsection{Novelty-driven Search}
\label{sec:exp}

Certain exploratory decisions can enhance the sample efficiency of a learning system by generating novel observations that help avoid locally optimal solutions that are biased by limited or incomplete datasets. There are two prominent methods that can be applied in this context~\cite{liu2023adventurer}.

Firstly, uncertainty-based methods~\cite{janz2019successor} can estimate the uncertainty, or variance, of the objective in reinforcement learning (RL) using a Bayesian posterior~\cite{liu2022farsighter}. By maximizing this uncertainty, the learning system can choose actions that provide more informative observations, thereby improving its learning efficiency. This approach allows the system to explore regions of the state-action space that may initially appear uncertain or unfamiliar, enabling the discovery of novel and potentially better solutions~\cite{halev2024microgrid}.

Secondly, intrinsic motivation methods~\cite{pathak2017curiosity} leverage the uncertainty of the next state as a reward signal to guide the learning system towards uncertain environments. By incorporating the uncertainty as a driving force, the controller actively seeks out and explores states that offer new and diverse experiences. This approach promotes exploration and helps the system navigate through complex search spaces, such as those encountered in farming scenarios, to discover innovative and effective solutions.

By intentionally diversifying its actions, such as adjusting the setpoints of heaters, in different states, a decision policy optimized for farming scenarios can apply similar approaches to explore and uncover novel solutions. These solutions not only contribute to navigating the complex search space but also reduce the data requirements for learning. By proactively exploring new possibilities, the system can discover effective strategies while minimizing the reliance on large amounts of data.

In summary, incorporating exploratory decisions that promote uncertainty and diversity can significantly enhance the sample efficiency of a learning system. By deliberately diversifying actions and exploring uncertain regions of the state-action space, the system can discover novel and effective solutions in complex environments, such as farming scenarios. These exploratory approaches not only improve learning efficiency but also reduce the data requirements for achieving optimal performance.
% take advantage of a well-designed exploration module because, as pointed out in~Sec.~\ref{sec:complex}, the state-action space may be extremely large and in~Sec.~\ref{sec:slow}, sampling can be time-consuming, and reward signals are sparse. 
% One potential limitation can be that the \emph{novel} state-actions pairs may violate \emph{safety} requirements, so a combined solution with constrained RL needs to be considered in practice~\cite{efroni2020exploration}. In the greenhouse task, we use RND~\cite{burda2018exploration}, a intrinsic-motivation exploration method, for policy searching in the leaned model and also help to request uncertainty samples from the greenhouse simulator server.

% the agent could learn a control policy with fewer samples. 
% However, the exploration may cause the agent to meet an unsafe state, e.g., a setpoint with a high temperature. Thus, we need to learn a controller under constraints, as we discussed in Sec~\ref{sec:pre} and Sec~\ref{sec:constrained}.

% \subsection{Constrained Optimization}
\subsection{Knowledge-guided Learning}
\label{sec:knowledge_guided_learning}

In this section, we suggest three different approaches that are aided by human expertise. 
In particular, some of those can be used to enable a farming system to not only maximize productivity but satisfy ``resource and safety constraints'' (Sec.~\ref{sec:constraints}) during optimization process. 
Moreover, another subset can mitigate other critical issues such as ``scarce data'' and/or ``delayed reward signals'', highlighted in~Sec.~\ref{sec:slow}. 

\subsubsection{Constrained RL}~\label{sec:constrained}
% For example, in the greenhouse controls, we should consider the temperature cannot exceed certain value.
% To consider the constraints in agricultural, we model the problem of learning with constraints as a Constrained Markov Decision Process (CMDP) as discussed in Sec~\ref{sec:pre}. 
% In a CMDP, the objective is to maximize long-term reward while keeping certain costs under their respective constraints. 
% In this paper, we divide constraints into two general types relevant to the CMDP problem: cumulative and instantaneous, defined as follows.
% \begin{itemize}
% \item Cumulative constraints require the sum or mean of one constraint variable from the beginning to the current time step to be within a certain limit (e.g., total usage of water, electricity, fertilizer, etc).
% \item Instantaneous constraints are constraints that the chosen action needs to satisfy in each step (e.g., temperature, Co2 capacity, etc in each loop step).  
% \end{itemize}

% ``Resource and safety constraints'' are often embedded in applications of RL to agriculture (cf.~Sec.~\ref{sec:constraints}). 
In Section \ref{sec:mdp}, we formalize the design of a Markov Decision Process (MDP) that incorporates two types of constraints based on domain knowledge: cumulative constraints and instantaneous constraints. Cumulative constraints involve factors such as the total use of gas, while instantaneous constraints relate to requirements like maintaining a minimum humidity level, as discussed in Liu et al. (2021)~\cite{liu2021resource}.

To address cumulative constraints, the key idea is to transform the constrained optimization problem into an unconstrained one. One common approach for handling such constraints is through Lagrangian relaxation, which was initially introduced in Altman (1999)~\cite{altman1999constrained}. By applying Lagrangian relaxation, the problem can be reformulated to optimize a modified objective function that accounts for the constraints~\cite{liu2020ipo}.

To satisfy instantaneous constraints, an effective method involves adjusting actions at each step by projecting them onto a feasible space for evaluation. This can be achieved, for example, by incorporating a projection layer at the end of the policy neural network~\cite{liu2021clara}. This projection layer ensures that the selected actions adhere to the instantaneous constraints, allowing for more reliable and feasible decision-making~\cite{liu2021cts2}.

Overall, the constrained RL approach provides an elegant framework for integrating practical concerns and constraints (as discussed in Section \ref{sec:constraints}) into the learning process of farming policies. By addressing cumulative and instantaneous constraints through techniques such as Lagrangian relaxation and action projection, the approach enhances the applicability and effectiveness of RL algorithms in real-world farming scenarios.
% In addition, evaluation results on the instantaneous criteria can play a role of ``dense reward signals'' for the trained policy at each time step whereas typical setups of plant system can only offer sparse signals after harvest, as discussed in Sec.~\ref{sec:slow}. 

% learn a greatly practical policy
% at a cost of computational resource and sample efficiency when the data collection is restricted with cumulative constraints. 
% In the greenhouse case, we set contraints for the temperature, $CO_2$ and light intensity etc, which guarantees all the setpoints value are significative in practise.
% In a given state, the unconstrained policy outputs an action and then passes it to the safety layer, which projects the action to the nearest feasible action.
% There is a trade-off between the constrained RL and efficient exploration. 
% The constrained RL with cumulative constraints would decrease the sample efficiency since the collected data is restricted, while efficient exploration encourages the agent explore more unseen state. 
% Moreover, the constrained RL would increase the computation complexity compared with typical RL.

% \begin{figure}
%     \centering
%     \includegraphics[width=0.9\linewidth]{figures/mbrl.pdf}
%     \caption{Model-based RL.}
%     % After training in a controlled environment, the learned model can be used to provide optimal treatment recommendations in the ﬁeld.}
%     \label{fig:mbrl}
% \end{figure}

\subsubsection{Learning from Human Demonstrations}~\label{sec:learn_from_demo}

% (causal modeling, modular (e.g., breakup climate and plant growth) so that simpler models. Not just a super-large NN.
    % RL can be particularly hard in farming systems, compared to other applications, because of their complexity (Sec.~\ref{sec:complex}), slowness (Sec.~\ref{sec:slow}), and safety requirements (Sec.~\ref{sec:constraints}). 
    % Another method to formally address the ``concerns of resource saving and safe operation'' (Sec.~\ref{sec:constraints}) can be 
    
To address the various constraints discussed in Section.~\ref{sec:constraints}, we recommend the adoption of a learning paradigm called imitation learning (IL), where a control policy is learned from demonstrations provided by human experts.

Specifically, the decision-making agent can be trained to mimic the past behaviors of human operators who have demonstrated state-action trajectories that adhere to all the relevant constraints. There are different approaches to implementing IL. One approach is to design the learning process to enforce the trained policy to replicate the exact decisions made by the demonstrators~\cite{yin2022cross}. Another approach is to adopt offline reinforcement learning (RL)~\cite{agarwal2020optimistic}, where demonstration data is used more loosely to estimate the values of taken actions by an RL algorithm, such as conservative Q-Learning~\cite{kumar2020conservative}.

These learning paradigms can also provide a solution to the ``lack-of-data problem" discussed in Section~\ref{sec:slow} if datasets of past farm operations are available. Additionally, IL can guide learning more effectively by computing ``denser rewards" compared to the sparse rewards commonly encountered in traditional RL (see Section~\ref{sec:slow}). This is achieved by assessing the similarity between the chosen actions of the agent and those of the demonstrator at each time step, enabling the learning process to provide more informative feedback.

As a specific example, the authors of a recent study\cite{tao2022optimizing} successfully applied IL to optimize a policy agent that simultaneously managed nitrogen levels and controlled irrigation in farming operations.

In summary, adopting IL as a learning paradigm allows control policies to be learned from demonstrations provided by human experts, offering a viable approach to address various constraints in agricultural decision-making. IL can leverage available datasets, provide dense rewards, and guide learning effectively, making it a promising solution for improving decision-making processes in farming systems.

\subsubsection{Reward Shaping}
\label{sec:reward_shaping}

% \subsubsection{Inverse RL}
% It is a major challenge for RL to process sparse and long-delayed rewards. It is difficult to untangle irrelevant information and credit the right action in each step. 
% In the greenhouse task, instead of only output the final net profit at the last step, we output the daily net profit at each step as the reward.
% % In real life, we establish intermediate goals for complex problems to give higher-quality feedback. 
% Nevertheless, such intermediate rewards are hard to establish for many RL problems.

% Inverse reinforcement learning (IRL) is the problem of inferring the reward function of an agent, given its policy or observed behavior acting in an environment.
% % as shown in Fig.~\ref{fig:irl}.
% We can fit a reward function with the use of professional demonstrations. Once a reward feature is fitted, we are able to use Policy Gradient, Model-based RL or different RL to locate the ideal policy. 
% For example, we are able to compute the policy gradient with the use of the reward feature as opposed to sampled rewards. With the policy gradient calculated, we optimize the policy closer to the finest rewards gain.

In cases where human demonstrations are unavailable, an alternative approach to obtaining denser feedback signals is through the technique of reward shaping. Reward shaping involves generating pseudo rewards based on domain knowledge, even before the completion of an episode, to encourage specific action selections over others~\cite{hu2020learning}.

The field of robotics has extensively explored and utilized reward shaping to enable robots to autonomously learn exploratory and purposeful actions towards achieving their ultimate goals~\cite{mezghani2022learning}. We propose adopting a similar approach in the context of farm management to generate more frequent assessments of the decisions made, even before reaching the harvest phase.

By leveraging domain expertise, relevant factors and metrics can be identified to shape the rewards in a way that guides the decision-making process. This approach allows the system to receive more informative feedback at intermediate stages of the farming process, promoting actions that align with the desired outcomes. For example, the system could be rewarded for maintaining optimal temperature and humidity levels or for efficiently utilizing resources such as water and fertilizers.

By incorporating reward shaping in farm management, decision-making algorithms can receive more frequent signals of assessment and guidance, facilitating the exploration of effective strategies throughout the entire farming process. This approach enhances the learning process and enables the system to make informed decisions that lead to improved outcomes, even before the harvest phase is reached.

% For example, as the controller selects a desirable decision according to the domain knowledge, the expert could offer some reward even though its positive consequence is still steps away.  
% Inverse reinforcement learning (inverse RL) considers the problem of extracting a reward function from observed (nearly) optimal behavior of an expert acting in an environment as shown in Fig.~\ref{fig:irl}.

% \begin{figure}[t]
%   \centering
%   \includegraphics[width=0.9\linewidth]{figures/IRL.png}
%   \caption{Inverse Reinforcement Learning.}
%   \label{fig:irl}
% \end{figure}

\subsection{Model Learning}

Learning from continuous interactions with a real-world farming environment can be risky (Sec.~\ref{sec:constraints}) and time-consuming (Sec.~\ref{sec:slow}), so a realistic replicate of environment can be used instead. 
This section explores it in the context to improve RL and briefly discusses the strategies to close the gap to the real complex environment (Sec.~\ref{sec:complex}).

\subsubsection{Model-based RL}~\label{sec:model}

%  In a standard RL setting, an environment, typically modeled by a MDP or CMDP, is given for the agent to explore and exploit by performing actions and then observing the state transitions and rewards. However, t
 
%  In real-world agricultural scenarios prevent the agent from easily accessing the environment.
%  For example, in the greenhouse control scenario, it is time and resource cost if the controller repeatably access the greenhouse.
 
% Since learning from continuous interactions with a real-world farming environment can be expensive (Sec.~\ref{sec:slow}),
% it is time consume and resource cost if the controller repeatedly accesses the farming system.

Model-based reinforcement learning (MBRL) is an approach that involves learning a model of the environmental dynamics to simulate realistic feedback, allowing a separate agent to interact with it~\cite{janner2019trust}. This learned environmental model can then be utilized by a typical RL algorithm, enabling rapid responses to the actions of the trained controller agent.

In the context of learning farm management, employing a model-based method can help address the challenges associated with the slow rate at which plant systems evolve (see Section~\ref{sec:slow}). Many previous studies have utilized simulators for agricultural tasks, although most of them were not built through a separate learning process but rather based on domain knowledge of the physical interactions among environmental factors~\cite{wu2022optimizing,cao2022igrow,tao2022optimizing,overweg2021cropgym}.

It is important to note that the accuracy of the modeled environment significantly impacts the subsequent performance of the trained agent~\cite{liu2025look}. In Section~\ref{sec:method}, we present a case study using this technique to delve into the problem in more detail.

By leveraging model-based RL, we can overcome the limitations posed by the slow dynamics of plant systems. The use of simulators based on domain knowledge, coupled with the learned environmental model, allows for efficient and effective training of the agent. This approach enables the agent to make informed decisions in a timely manner, contributing to improved farm management practices.

\subsubsection{Sim2Real}
% Previous works have empirically shown signiﬁcant sample eﬃciency improvements. 
Another significant challenge in applying simulation-based methods to farm management is the simulation-to-real gap, where the simulated feedback does not perfectly replicate the complexities of the real-world environment (see Section~\ref{sec:complex}). Consequently, controller parameters optimized on the learned model may not generalize well to real-world scenarios.

To address this performance gap, several approaches can be considered:

1)~\emph{System identification} to build a reliable simulator based on mathematical, physics-based models for a real agricultural system;
2)~\emph{Domain adaptation} to shape the data distributions or representations from a simulator to match those in real scenarios~\cite{chen2021cross}; and  
3)~\emph{Domain randomization} to create random variants of simulated environment so that this augmented distribution could include plausible observations from real environments~\cite{lee2019network}.

While these approaches have been explored in other domains, their application to farm management is relatively unexplored. However, considering the potential impact, further research in this direction is warranted. By addressing the simulation-to-real gap, we can enhance the effectiveness and practicality of simulation-based methods for farm management, ultimately improving decision-making processes and outcomes in real-world agricultural settings.

% –a Digital Twin basically is a digital equivalent of a real-life object of which it mirrors its behaviour and its states over its lifetime in a virtual space
% especially the management aspects of using Digital Twins to plan, monitor, control and optimize farm processes need to be further studied.

\subsection{Learning to Learn}

In general, agricultural systems are too complex with diversity for a decision policy trained on an environment to ensure its perfect generalization to another (Sec.~\ref{sec:complex}). 
We here discuss two useful methods for efficient learning possibly for a new task on a novel environment as a solution to the ``data scarcity'' (Sec.~\ref{sec:slow}).

\subsubsection{Meta Learning}~\label{sec:meta}
Meta-learning, as a powerful approach, offers the ability to assess the suitability of a policy across a wide range of learning environments. By leveraging this meta-information, we can initialize the learning process in new environments to achieve optimal performance in the fastest possible manner ~\cite{vanschoren2019meta}. This concept has also been applied in combination with reinforcement learning (RL) to develop policies that require only a short learning session to effectively adapt to previously unseen environments~\cite{duan2016rl}.

In the context of farm management, depending on the specific farm configuration, such as the types of crops and controllable facilities, we can employ meta-learning to train an optimal learner for farm automation. By leveraging meta-information, the learner can quickly adapt and perform effectively in diverse farm environments.

By integrating meta-learning with RL techniques, we can significantly reduce the time and effort required to train policies for farm automation. This approach holds great promise for enabling efficient and adaptive decision-making in agricultural settings.

\subsubsection{Transfer Learning}~\label{sec:tl}

Unlike meta-learning, transfer learning (TL) involves fine-tuning an already trained controller using new data from a different environment. This approach has gained significant attention in recent years, particularly in the field of reinforcement learning (RL), as it allows for improved performance with a smaller number of samples~\cite{zhu2020transfer}.

In the context of farm management learning, transfer learning can be a highly effective approach, especially considering the high cost associated with data collection (as discussed in Sec.~\ref{sec:slow}). However, it is worth noting that transfer learning in the domain of farm management has received relatively little attention thus far, despite its potential to address the data scarcity issue and improve learning efficiency.

By leveraging transfer learning techniques, we can leverage knowledge gained from previous training to accelerate the learning process in new environments. This can lead to more efficient and effective farm management systems that require fewer data samples for training, ultimately improving sustainability and productivity in agriculture.

\subsection{Explainable Decision-Making}~\label{sec:explain}

As highlighted in Sec.~\ref{sec:interact_with_human}, agriculture involves human interaction, even though it is moving towards automation. Thus, exploring the adaptation of ``explainable AI" from other domains to farming represents a highly impactful avenue for future research.

In the realm of explainable RL (XRL), different approaches can be categorized based on the elements they aim to explain~\cite{milani2022survey}. These categories include:

$1$)~salient features that affect the choice of action~$a_t$ for input state~$s_t$, 
$2$)~the MDP to highlight particular experiences that has led to the current behavior, and 
$3$)~long-term policy-level behaviors to evaluate its overall capacity.

Any of these explanation techniques can significantly enhance the interaction between AI farm managers and human farmers, fostering trust in the generated decisions. Consequently, extending existing frameworks from the list of works in Table~\ref{tab:list_of_work} with explainability capabilities holds promise as a future research direction.

% PL explanations present summaries of long-term behavior through abstraction or representative examples.
% which are critical for understanding a controller’s behavior to evaluate its overall competency.

% Our model for the greenhouse as shown in Fig.~\ref{fig:model}, can explain the relations of outside climate and indoor climate, as well as the the relations of the plants and indoor climate.

% \section{Community action items}

% \subsection{publicly available dataset}
% \subsection{build a simulation environment (OpenAI Gym for agriculture) Many subfieds, different crops, diff. applications, with fixed application interface to allow people to adapt. }

\subsection{Multimodal ML}~\label{sec:multi}

The state of crops can be captured through various sensors, generating data in different formats such as visual images and time-series climate or chemical measurements. Additionally, human expert knowledge can provide valuable insights into the observed facts and their relationships within the plant system. To leverage the full potential of these diverse data sources, we propose the adoption of "multimodal" machine learning and reinforcement learning techniques~\cite{baltruvsaitis2018multimodal,ma2022multimodal}.

In multimodal ML/RL, decision-making is based on states that can be composed of combined forms of data, including visual, auditory, or text inputs. This approach allows for a more comprehensive understanding of the environment. Furthermore, integrating human expertise into the learned policy can lead to adaptive decision-making (see Sec.~\ref{sec:interact_with_human}). A similar approach was demonstrated by~\cite{silva2019neural}, where human experts provided decision trees for gaming environments, enabling the RL agent to initialize its learning process and achieve better performance.

Applying multimodal techniques and incorporating human expertise into the learning process has the potential to enhance decision-making in farming applications, leading to improved agricultural management.

\section{Problems, Strategies, \& Lessons Learned from autonomous Greenhouse Control}
\label{sec:method}

\begin{table}
\centering
\begin{adjustbox}{width=\columnwidth}
\begin{tabular}{|c | c c c c c|} 
 \hline
 Rank & $1$ & $2$ & $3$ & $4$ & $5$ \\ [0.5ex] 
 \hline\hline
 Team & Koala & \textbf{AggieAI} & LetTUce Win! & IUACAAS & OnePlanet \\ 
 \hline
 Net Profit & $8.68$ & $\bm{8.00}$ & $7.96$ & $7.63$ & $7.51$ \\
 \hline
\end{tabular}
\end{adjustbox}
\caption{Public board of the $3$rd Autonomous Greenhouse Challenge, where our record is shown to be bold in the second place.} 
\label{tab:scores}
\end{table}

As in literature in Table~\ref{tab:list_of_work}, our team first considered using RL combining with BO~algorithms to gain an optimal control policy from repeated interactions with the simulator. 
We, however, encountered many challenges described in Sec.~\ref{sec:challenges}---first, ``data scarcity'' (Sec.~\ref{sec:slow}) primarily because each of four greenhouse environments on simulation could be accessed only for a limited number of times per day according to the rules of competition. For instance, the last environment (Simulator~D), which was used for the final team assessment, would be available only for an hour, and each team could run it only $200$~times.
Although other configurations such as Simulator~A could be queried $1,000$~times a day, it would also be uncertain whether a trained model there could ``generalize'' well to Simulator~D, as discussed in Section~\ref{sec:complex}. 
Furthermore, the growth cycle of tens of days would be a too long time scale for each hourly decision to be properly evaluated by the eventual net profit computed at the end---i.e.,~problem of ``delayed rewards'', explained also in Section~\ref{sec:slow}. 

% \subsubsection{Model-based RL} \label{sec:mbrl}
% \begin{figure*}[t]
%     \centering
%     \includegraphics[width=0.8\linewidth]{figures/model.pdf}
%     \caption{Greenhouse model.}
%     % After training in a controlled environment, the learned model can be used to provide optimal treatment recommendations in the ﬁeld.}
%     \label{fig:model}
% \end{figure*}

% \begin{figure}[t!]
%      \centering
%      \begin{subfigure}[t]{0.2\textwidth}
%          \centering
%          \includegraphics[width=0.2\textwidth]{figures/temp1.png}
%          \caption{$y=x$}
%          \label{fig:y equals x}
%      \end{subfigure}
%      % \hfill
%      \begin{subfigure}[t]{0.2\textwidth}
%          \centering
%          \includegraphics[width=0.2\textwidth]{figures/temp2.png}
%          \caption{$y=3\sin x$}
%          \label{fig:three sin x}
%      \end{subfigure}
%      % \hfill
%      \begin{subfigure}[t]{0.3\textwidth}
%          \centering
%          \includegraphics[width=0.2\textwidth]{figures/plant_head1.png}
%          \caption{$y=5/x$}
%          \label{fig:five over x}
%      \end{subfigure}
%         \caption{Three simple graphs}
%         \label{fig:three graphs}
% \end{figure}
\begin{figure}[t]%
\centering
\subfigure[][]{%
\label{fig:pred_temp1}%
\includegraphics[height=1.1in]{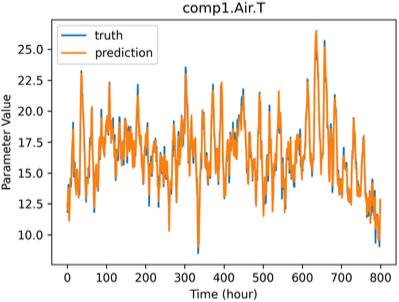}}%
\hspace{8pt}%
\subfigure[][]{%
\label{fig:pred_temp2}%
\includegraphics[height=1.1in]{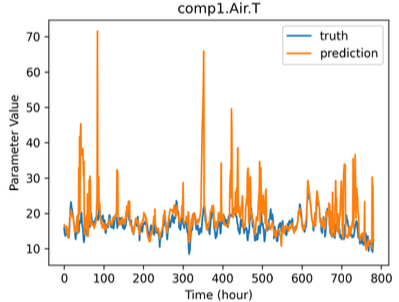}} \\
\subfigure[][]{%
\label{fig:pred_plant1}%
\includegraphics[height=1.1in]{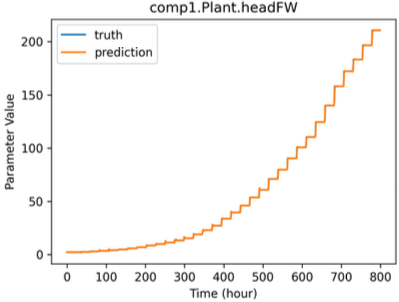}}%
\hspace{8pt}%
\subfigure[][]{%
\label{fig:pred_plant2}%
\includegraphics[height=1.1in]{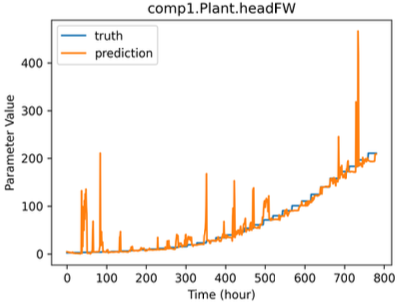}}%
\caption{Prediction examples in our MBRL method:
\subref{fig:pred_temp1} and \subref{fig:pred_temp2} are the $1$- and $20$-step lookahead predictions of indoor temperature, respectively while \subref{fig:pred_plant1} and \subref{fig:pred_plant2} are those of fresh weight of lettuce.}%
\label{fig:pred_temp_plant}%
% \vspace{-0.3cm}
\end{figure}

To address some of those issues, we explored a variant of RL approach---model-based RL~(MBRL), which is to first model the environmental dynamics from observations so the subsequent session of typical policy learning could take advantage of the generated data under the model, even though the target environment has been inaccessible. 
Building an accurate model is, hence, essential, in which its state would evolve as it would in the original environment. 

% For example, in the greenhouse control task, it is time and resource cost if the controller repeatably access the greenhouse server.
We deploy supervised model to approximate the greenhouse. 
Given current state input $s_t, a_t$, the model outputs the reward $r_t$ and next state $s_{t+1}$. 
We assume the plant statistics is only affected by previous plant statistics and current indoor climate which is affect by the control setpoints, previous indoor climate and outdoor climate.
Specifically, we built two models for the climate and plant separately. The climate model decides the indoor climate based on the input of control setpoints, previous indoor climate and outdoor climate. The plant model decides the plant statistics based on the input of indoor climate and previous plant statistics.

Figure~\ref{fig:pred_temp_plant} shows some of the predictive results from a feed-forward neural network that has been trained to estimate the temperature or lettuce' fresh weight data various steps ahead based on the latest state information. 
As the size of prediction window increases from~$1$ to~$20$, overall accuracy generally \emph{declines} due to the accumulated errors from the predictions at intermediate steps. 
This observation basically supports our claim of ``system complexity'' as another challenge in research (Section~\ref{sec:complex}).

Here, \emph{we thus claim that MBRL would be more useful in scenarios where each episode tends to involve a smaller number of steps.} 
For our case study, we decided \emph{not} to adopt MBRL, since an episode could be around $1,000$~steps long, which would lead to much worse predictions. 
A more powerful network could have offered a better result, but we have not investigated it further for the sake of time during competition. 
\begin{table}
\centering
% \scriptsize
\begin{adjustbox}{width=\columnwidth}
\begin{tabular}{| l | c c c c|} 
 \hline
 Parameter & A & B & C & D \\ [0.5ex] 
 \hline\hline
 num\_days & $38$ & $42$ & $40$ & $37$ \\ 
 heatingTemp\_night & $10$ & $5$ & $3$ & $6$ \\
 heatingTemp\_day & $26$ & $20$ & $20$ & $19.5$ \\
 CO$2$\_purCap & $182$ & $184$ & $270$ & $245$ \\
 CO$2$\_setpoint\_night & $694$ & $488$ & $400$ & $545$ \\
 CO$2$\_setpoint\_day & $1200$ & $955$ & $1200$ & $1130$ \\
 CO$2$\_setpoint\_lamp & $912$ & $990$ & $0$ & $1100$ \\
 light\_intensity & $0$ & $64$ & $0$ & $3$ \\
 light\_hours & $12$ & $6$ & $9$ & $8.6$ \\
 light\_endTime & $24$ & $20$ & $17$ & $17$ \\
 light\_maxIglob & $300$ & $267$ & $800$ & $800$ \\
 scrl\_ToutMax & $8$ & $8$ & $5$ & $4.8$ \\
 vent\_startWnd & $50$ & $50$ & $50$ & $51.7$ \\
 \hline\hline
 NetProfit & $4.941$ & $4.714$ & $7.966$ & $\bm{8.003}$ \\
 \hline
\end{tabular}
\end{adjustbox}
\caption{Optimal values of control parameters that our BO~algorithm discovered for four different greenhouse simulations. The net profit achieved at Simulator~D was used for assessment in Table~\ref{tab:scores}.} 
\label{tab:opt_params}
% \vspace{-0.4cm}
\end{table}

\subsection{Bayesian Optimization}
To efficiently search the available parameter space of control policy, our team chose to utilize a ``BO''~algorithm (Sec.~\ref{sec:bo}), as shown in Fig.~\ref{fig:bo}. 
More specifically, the implementation of a Python library, so-called Hyperopt\footnote{http://hyperopt.github.io/hyperopt}, was used to employ the Tree-structured Parzen Estimator~(TPE) algorithm~\cite{bergstra2011algorithms}.

As a preliminary study on BO, we designed a relatively simplistic method in which the optimal values of most parameters would be applied throughout the simulated growing cycle (around $40$~days). 
Yet, we also selected such parameters as heating temperature and $\text{CO}_2$~dosing to adjust to different values at the times of ``sunrise'' and ``sunset'' each day.  

Table~\ref{tab:opt_params} offers the list of the specific selection of parameters optimized to maximize the outcome of net profit. 
\emph{Surprisingly}, this simple method led to a highly successful result, which contained the second highest net profit at the competition particularly on Simulator~D. 
From this finding, \emph{we have learned that BO is a highly promising method because ($1$) it could optimize a dozen of parameters simultaneously even from a limited number of samples, and ($2$) improvement would be expected, if a more sophisticated approach was employed to use optimal decisions updated more frequently.}  

One limitation we have seen is that BO~algorithms can ``slow down'' significantly, as the number of drawn samples increases. 
In our setting, for instance, a noticeable increase in running time was observed particularly when over $200$~samples were used for learning. 
The bottleneck here is at the process that models those samples with a multivariate Gaussian distribution, and it would be worsened if more parameters are considered for optimization. 
Thus, \emph{if optimization still needs to be scaled up, we suggest utilizing libraries like BoTorch\footnote{\url{https://botorch.org/}}, which enables the feature of GPU~acceleration to boost up the computational speed during~BO.}

\section{Conclusion}
This survey paper has offered a broad overview of existing approaches to building fully autonomous decision-making systems for farm management. 
In particular, we have categorized those into six groups depending on the task studied and also identified the most widely adopted methods to learn effective control policies from agricultural data. 
Moreover, we have discussed general challenges and promising solutions in such research, and our success in the $3$rd Autonomous Greenhouse Challenge has also been introduced with specific lessons learned. 

% \appendix
% \section{Reference Examples}

% \section{Acknowledgments}
% The work was partially supported through grant USDA/NIFA 2020-67021-32855, and by NSF through  IIS-1838207, CNS 1901218, OIA-2134901.

\bibliography{aaai23}

\end{document}